\documentclass[lettersize,journal]{IEEEtran}
\usepackage{amsmath,amsfonts}
\usepackage{algorithmic}
\usepackage{algorithm}
\usepackage{array}
\usepackage[caption=false,font=normalsize,labelfont=sf,textfont=sf]{subfig}
\usepackage{textcomp}
\usepackage{stfloats}
\usepackage{url}
\usepackage{verbatim}
\usepackage{graphicx}
\usepackage{cite}
\usepackage{indentfirst}
\usepackage{CJK}
\usepackage{threeparttable}
\usepackage{multirow}
\usepackage{graphicx}
\usepackage{algorithm,algorithmic}
\usepackage{bbding}
\usepackage{mathtools}
\usepackage{xcolor}
\usepackage{amsmath}
\usepackage{comment}
\usepackage{stackengine}
\newcommand{\etal}{\textit{et al}.}
\DeclareMathOperator*{\argmax}{arg\,max}

\begin{document}
\renewcommand{\algorithmicrequire}{\textbf{Input:}}  
\renewcommand{\algorithmicensure}{\textbf{Output:}} 

\title{Progressive with Purpose: Guiding Progressive Inpainting DNNs through Context and Structure}

\author{Kangdi Shi, Muhammad Alrabeiah, Jun Chen

\thanks{Kangdi Shi, Jun Chen (corresponding author) are with the Department of Electrical and Computer Engineering, McMaster University, Hamilton, ON L8S 4L8, Canada.(email: shik9@mcmaster.ca; junchen@mcmaster.ca)}

\thanks{Muhammad Alrabeiah is with the Electrical Engineering Department, King Saud University, Riyadh 11451, Saudi Arabia. (email: marrabeiah@ksu.edu.sa)}}

\maketitle

\begin{abstract}
The advent of deep learning in the past decade has significantly helped advance image inpainting. Although achieving promising performance, deep learning-based inpainting algorithms still struggle from the distortion caused by the fusion of structural and contextual features, which are commonly obtained from, respectively, deep and shallow layers of a convolutional encoder. Motivated by this observation, we propose a novel progressive inpainting network that maintains the structural and contextual integrity of a processed image. More specifically, inspired by the Gaussian and Laplacian pyramids, the core of the proposed network is a feature extraction module named GLE. Stacking GLE modules enables the network to extract image features from different image frequency components. This ability is important to maintain structural and contextual integrity, for high frequency components correspond to structural information while low frequency components correspond to contextual information. The proposed network utilizes the GLE features to progressively fill in missing regions in a corrupted image in an iterative manner. Our benchmarking experiments demonstrate that the proposed method achieves clear improvement in performance over many state-of-the-art inpainting algorithms.
\end{abstract}


\section{Introduction}
\label{sec:introduction}
\IEEEPARstart{I}{mage} inpainting is the task of restoring missing patches of pixels in an image \cite{Jireh2021,bertalmio2000image}. As the name suggests, inpainting targets filling in missing parts of an image (i.e., image holes) with contextually meaningful information so that the image can be restored to its original form. This could be a quite difficult task for machines, for it is an ill-posed inverse problem \cite{Guillemot2014}. It requires not only the ability to predict what is missing, but also whether it fits within the context of the image or not. Thus, a key to attaining satisfying inpainting results is to ensure that the reconstructed pixels are consistent with the uncorrupted region and exhibit coherence in both structure and texture.

As a main remedy for restoring image quality, inpainting is of great importance nowadays, for modern societies are increasingly reliant on visual content with images as its building block, from surveillance systems to autonomous vehicles, media streaming, and conference calls. Storing, displaying, and exchanging huge amount of images make them prone to damages, one of which is missing pixels (image holes). It is thus unsurprising to see increasing research interest in inpainting within the computer vision community.

\subsection{Motivation}
Many image inpainting techniques have been proposed over the past two decades. They could loosely be grouped into two major categories: traditional and modern. The main defining difference between the two categories is the use of deep learning. The traditional category of techniques could collectively be divided into two sub-categories \cite{Guillemot2014}: exemplar-based and diffusion-based. The former approach\cite{barnes2009patchmatch} searches for the best matching patches from known regions and pastes them into missing regions. Such techniques have high computational costs for patch searching and generate unrealistic results due to the lack of perspective transformation. The diffusion-based techniques, on the other hand, recreate a missing region with features from its surrounding known region. Although diffusion-based techniques are more efficient than their exemplar-based counterparts, they result in over-smoothed inpainting results because of regularization based on partial-differential equations.

The advent of deep learning in computer vision has created a surge of inpainting techniques that utilize Deep Neural Networks (DNNs). Although those techniques exhibit some overlap with the tradition diffusion-based techniques, they define the state-of-the-art in inpainting, and, therefore, merit a category of their own. 
Most of the early works on inpainting with deep learning, like\cite{ren2019structureflow}\cite{nazeri2019edgeconnect}, follow a two-stage approach, which firstly learns the image structure from a given edge/structure map of a corrupted image, then refines the missing region with a texture generator. However, two-stage image inpainting methods usually cause artifacts due to their limited ability to recover both structure and texture.

To deal with those artifacts, progressive inpainting techniques have been explored. They rely on the idea that not all predicted pixels in a region plagued with artifacts are defective; some are good predictions that could be utilized to improve the re-generated region and weed out the artifacts. Hence, those techniques fill in the missing holes by iterating over the image and learning from previously-predicted pixels. Examples of such techniques are the full-resolution residual network proposed by Guo \etal\cite{guo2019progressive} and the iterative confidence feedback network proposed by Zheng \etal\cite{zeng2020high}. Despite the improvement they provide over two-stage techniques, the performance of progressive techniques is still prone to artifacts. This could be traced back to the inexplicit modelling of structure and texture in those techniques.

Fusing structure and texture awareness with progressive inpainting is arguably the most promising approach to overcoming visual artifacts, which shall be followed in this paper. Developing structure and texture-aware algorithms has been explored recently in Guo \etal\cite{guo2021image}. Two different but coupled autoencoders are trained with structure and texture constraints to fill in holes in corrupted images. The results are encouraging, but the algorithm can cause distortion in deep parts of the hole due to one-stage feature fusion. This could be overcome by incorporating progressive inpainting into the learning process.

\subsection{Contribution}
In an attempt to bring together progressive learning and fusion of texture and structure, this paper presents a Gaussian-Laplacian feature Extraction (GLE) module. The main contributions of the proposed architecture are summarized below:

\begin{itemize}
	\item[$\bullet$]
	\textbf{GLE Module:} Inspired by image pyramid, we propose a GLE module to obtain features from high and low image frequency components. Those components provide texture information (low-frequency components) and structure information (high-frequency components). The GLE module leverages those multi-frequency components to learn textural and structural features.
\end{itemize}
\begin{itemize}
	\item[$\bullet$]
	\textbf{Iterative Reinpainting Component:} A progressive reinpainting component is developed such that it gradually fills in the corrupted regions of an image. It utilizes features learned by the GLE modules from different frequency components to fill in the outer edge of the corrupted regions iteratively until the region is restored.
\end{itemize}
\begin{itemize}
	\item[$\bullet$]
	\textbf{Benchmarking and Evaluation Experiments:} Various experiments are designed to evaluate the performance of the proposed architecture and show the benefits of the GLE module and the reinpainting component. The experiments also compare the proposed inpainting algorithm to some state-of-the-art algorithms to situate its contribution to the inpainting problem.
\end{itemize}

\subsection{Paper organization}
The organization of this paper is shown as follows. Section II reviews works that are related to our method. Section III details the architecture of the proposed progressive image inpainting network. Sections IV and V illustrate the experimental setup and the experimental results. Section VI concludes this paper.

\section{Related Work}
The proposed solution is developed on top of a rich literature of image inpainting with deep learning. To facilitate the discussion, the following three subsections will review some concepts related to the proposed solution and some relevant inpainting solutions. They should lay the necessary groundwork for the detailed description in Section \ref{sec:prop_sol}

\subsection{Variants of Texture and Structure Inpainting}
Inpainting based on texture and structure has been attempted in various forms in the literature. The concepts of \textit{style} and \textit{content} have been introduced in \cite{liao2019artist}, which could be viewed as derivatives of texture and structures, respectively. They are used to build a two-stage inpainting DNN. In the first, two encoders extract style and content latent information separately, and the second stage synthesize a full image from that information. Semantic segmentation masks are another alternative that helps capture structure information. They have been utilized in \cite{liao2020guidance, liao2021image, liao2020uncertainty} as a way to guide texture generation. In all those papers, an encoder network learns to generate a latent representation of the corrupted image that captures the structure. It does so by pushing the decoder network to recover not only the inpaintd image but also its segmentation mask. The three papers differ in the details of how to encode a corrupted image and generate a structurally consistent image, but they all match in the objective, capturing the structure to produce meaningfully inpainted image.

\subsection{Progressive Image Inpainting}
Progressive image inpainting, as the name suggests, aims to recover images gradually by utilizing features from undamaged and recently recovered regions. Overall, algorithms following this approach could be grouped in two broad categories: (i) contextual information-based algorithms, and (ii) structural constraints-based algorithm. Both are briefly reviewed below.

Contextual information-based algorithms rely mainly on CNN features extracted from input images to restore damaged regions. As a pioneer of the contextual features-based algorithms, Hsu \etal\cite{hsu2017high} propose using several deep convolution networks to learn progressive inpainting from multiple image scales, from low to high resolution images. Zhang \etal\cite{zhang2018semantic} recognize how inpainting lends itself to recurrent modelling; they propose to use several generative networks inter-connected with LSTM module, which progressively fills in the missing region of an image. More recently, Li \etal\cite{li2020recurrent} extend that analogy further. They propose a recurrent feature reasoning module with knowledge-consistent attention, which can progressively enhance the details in masked regions.

Compared with their contextual information-based counterparts, the structural constraints-based algorithms take advantage of  additional external structural constraints provided by edge detection algorithms. The algorithms in \cite{xiong2019foreground}\cite{nazeri2019edgeconnect} utilize contour or edge maps as a guide for image completion. To progressively complete the image, Li \etal\cite{li2019progressive} propose a U-net that recovers the edge maps while inpainting images progressively. These approaches, collectively, seek to tackle image inpainting by introducing structural constraints, yet their performance remains limited by a lack of information for recovering deeper pixels in the missing regions.

\subsection{Gaussian and Laplacian Pyramid}
A classical approach to image inpainting is centered around the idea of building multi-scale image pyramids, in which inpainting is done progressively from one scale to another---commonly from smallest to largest scale. Those pyramids are usually called Gaussian or Laplacian pyramids based on the type of filters used to generate them. Specifically, let $\mathcal{G}$ denote the Gaussian smooth operator, $\mathbf{I}_{\tau}$ express the input image to the ${\tau}_{th}$ level of Gaussian pyramid. $\mathcal{Q}$ denotes upsmpling operation, $\mathcal{D}$ denotes downsampling operation. The formulas for the output images $\mathbf{G}_{\tau}$, $\mathbf{F}_{\tau}$ of ${\tau}_{th}$ Gaussian pyramid and Laplacian pyramid are 
\begin{equation}
	\mathbf{G}_{\tau} = \mathcal{D}(\mathcal{G}(\mathbf{I}_{\tau}))
\end{equation}
\begin{equation}
	\mathbf{F}_{\tau} = \mathbf{I}_{\tau} - \mathcal{Q}(G_{\tau})
\end{equation}
Inpainting algorithms using Gaussian and Laplacian pyramids could roughly be clustered into three groups: inpainting on multiple Gaussian pyramids\cite{farid2010image}, inpainting on multiple Laplacian pyramids pyramids\cite{lee2016laplacian}\cite{padmavathi2014laplacian}, and inpainting on multiple Gaussian and Laplacian \cite{padmavathi2014laplacian}.

The difference between the first and the second group is in how the inpainting algorithm is applied on different image pyramids. For instance, Farid \etal\cite{farid2010image} first generate multiple Gaussian pyramids until most missing pixels are eliminated by the smoothing operation. Then, their algorithm copies and pastes the missing pixels from the small-scale image (top of the pyramid) to the large-scale images (bottom of the pyramid). In contrast, \cite{lee2016laplacian} utilizes Laplacian pyramid with patch search to recover missing pixels from small to large scale images in the pyramid. Because of the limitation of exemplar-based methods, both kinds of methods suffer from unrealistic inpainting results.

Benefiting from the combination of structure and texture, the third inpainting group (i.e., algorithms relying on multiple Gaussian and Laplacian pyramids) usually achieve better performance than their counterparts relying only on one of the two pyramids. However, the additional cost from inpainting both pyramids is relatively high compared to that of the former two groups.

\section{Proposed Inpainting Algorithm}\label{sec:prop_sol}
Like a person solving a jigsaw puzzle, an inpainting algorithm should fill in the missing regions by gradually piecing pixels together while keeping an eye on context and structure. Progressive algorithms, as mentioned earlier, restore missing pixels gradually using undamaged and recently recovered pixels, yet they do not jointly maintain contextual and structural information. This observation fuels the work in this paper; a Deep neural Network (DNN) is designed such that it progressively inpaints with the purpose of maintaining structure and context information. Hence, it is described as being progressive with purpose.

\begin{figure*}
	\includegraphics[width=\linewidth]{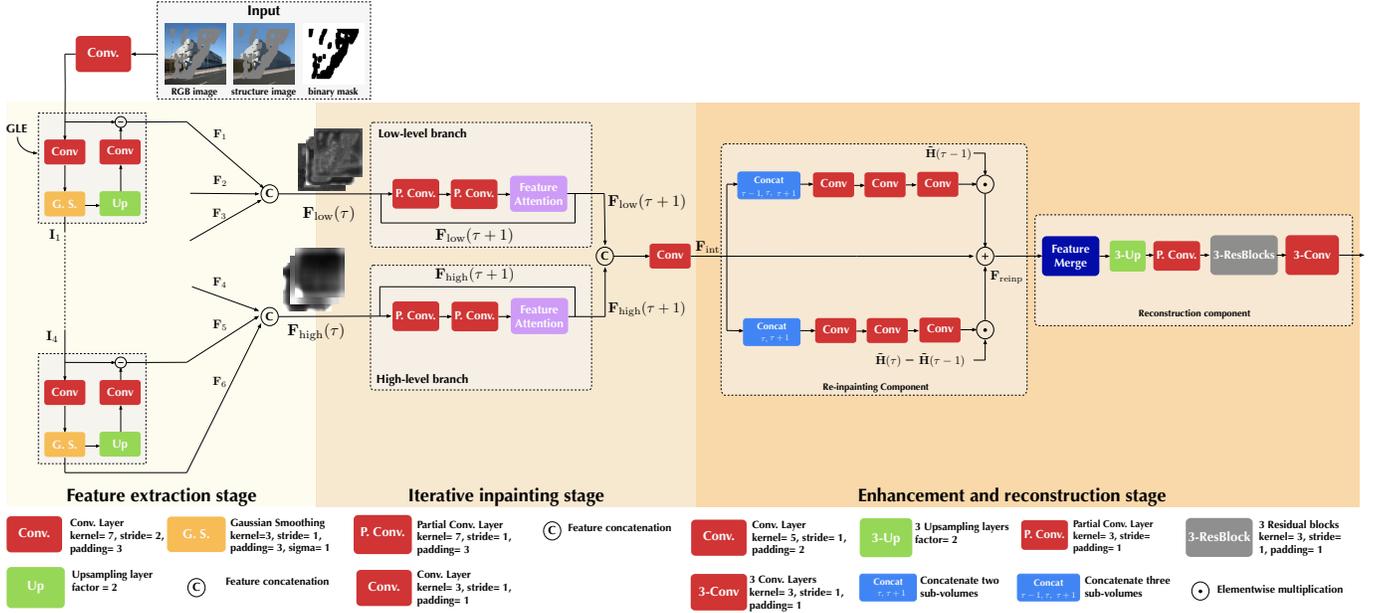}
	\caption{A graphical description of the proposed solution. It shows all three stages and details their main components and elements. In feature extraction stage, all input images are converted into 6 feature volumes: $\mathbf{F}_1$, $\mathbf{F}_2$, $\mathbf{F}_3$, $\mathbf{F}_4$, $\mathbf{F}_5$, $\mathbf{F}_6$ which are classified as low and high-frequency feature volumes: $\mathbf{F}_{\text{low}}(0)$ and $\mathbf{F}_{\text{high}}(0)$. In iterative inpainting stage, these feature volumes ( $\mathbf{F}_{\text{low}}(\tau)$ and $\mathbf{F}_{\text{high}}(\tau)$ $(0 \leq \tau \le 6)$) are utilized to recover the missing region progressively and generate inpainted features from each iteration, namely $\mathbf{F}_{\text{int}}(\tau)$. Furthermore, the re-inpainting component enhances each inpainted feature by leveraging features from neighboring iterations, and it stores enhanced features as $\mathbf{F}_{\text{reinp}}(\tau)$. In the end, the reconstruction component uses all enhanced features to produce the fully recovered image.}
	\label{fig:Framework}
\end{figure*}

The idea behind the proposed algorithm is to break down the inpainting task into three main stages, namely feature extraction (first stage), iterative inpainting (second stage), and enhancing and reconstruction (third stage). The first stage is aimed to extract multi-level features from the corrupted image, which mimics, to some extent, feature extraction from image pyramids used in classical inpainting algorithms such as \cite{farid2010image, lee2016laplacian}. The multi-level features capture contextual and structural information. They are fed to the iterative inpainting stage, which attempts to recover some of the missing information gradually over several iterations. Each one generates a pair of feature volumes. The pairs are passed to the enhancement and reconstruction stage to enhance the recovered information, fuse them into one feature volume, and reconstruct the complete image. The architecture of the proposed algorithm is depicted in Figure \ref{fig:Framework}.

The architecture is detailed in the following four subsections. The first one presents a formal description of how progressive inpainting restores missing pixels. The following three are a deep-dive into the three stages of the proposed architecture, describing the inner workings of each stage. Finally, the last subsection presents the loss function used to train the architecture.
\subsection{Rationale Behind the Proposed Algorithm}
Let $\mathbf{a}$ be the original image, and $\mathbf{b}$ be the  corrupted image.  We denote   the conditional probability distribution of the original image given  the corrupted image by $p_{\mathbf{A}|\mathbf{B}}$. Image inpainting can be formulated as a maximum a posterior (MAP) estimation problem: 
\begin{equation}
	\hat{\mathbf{a}}_{map} = \underset{\hat{\mathbf{a}} }\argmax \hspace{0.5mm} p_{\mathbf{A}|\mathbf{B}} (\hat{\mathbf{a}} | \mathbf{b}).
\end{equation}
Let $\mathbf{m}$ be the ground truth of the corrupted region, $\mathbf{n}$ and $\mathbf{c}$ be the valid region and corrupted region of the corrupted image, then $\mathbf{a} = \mathbf{m} \cup \mathbf{n} $, $\mathbf{b} = \mathbf{c} \cup  \mathbf{n}$. 
Note that   the conditional distribution of $\mathbf{m}$ given $\mathbf{n}$, denoted by
$p_{\mathbf{M} | \mathbf{N}}$, is a projected version of $p_{\mathbf{A} | \mathbf{B}}$ and can be learned from the training dataset. The MAP estimation of $\mathbf{a}$ based on $\mathbf{b}$ can be reduced to the MAP estimation of $\mathbf{m}$ based on $\mathbf{n}$:
\begin{equation}
	\hat{\mathbf{m}}_{map}= \underset{\hat{\mathbf{m}} }\argmax \hspace{0.5mm} p_{\mathbf{M} | \mathbf{N}} (\hat{\mathbf{m}} | \mathbf{n}).
\end{equation}
Clearly,  we have
\begin{equation}
	\hat{\mathbf{a}}_{map} = \hat{\mathbf{m}}_{map} \cup \mathbf{n}.
\end{equation}
Our algorithm aims to produce an approximate version of $\hat{\mathbf{m}}_{map}$.

We divide the corrupted region into $T$ concentric regions, and progressively recover the corrupted region in an inward manner from the $1_{th}$ to the ${T}_{th}$ concentric region.

Let  $\tilde{\mathbf{m}}_{r}^{(\tau)}$ denote the inpainted $r_{th}$ concentric region  at the $\tau_{th}$ step. The process proceeds as follows.
At the $\tau_{th}$ step (with $\tau$ from $1$ to $T$), we generate $\tilde{\mathbf{m}}_{\tau}^{(\tau)}$ based on the valid region $\mathbf{n}$ and the $\tau-1$ inpainted concentric regions $\tilde{\mathbf{m}}_r^{(\tau-1)} (1 \leq r \leq\tau-1)$ from the $\tau-1_{th}$ step\footnote{when $\tau=1$, there is no inpainted region available, but only the valid region $\mathbf{n}$.}, then refine $\tilde{\mathbf{m}}_r^{(\tau-1)} (1 \leq r \leq \tau-1)$ to $\tilde{\mathbf{m}}_r^{(\tau)} (1 \leq r \leq \tau-1)$, respectively. 

At the end of the $T_{th}$ step, we collected  $\tilde{\mathbf{m}}_{r}^{(\tau)}$($1 \leq \tau \leq T$, $1 \leq r \leq \tau$) generated throughout the process and perform an enhancement. Specifically, for $\tau$ from $1$ to $T-1$, we leverage $\cup_{r=1}^{\tau-1}\tilde{\mathbf{m}}_r^{(\tau-1)}$ (which is void when $\tau=1$) and $\cup_{r=1}^{\tau}\tilde{\mathbf{m}}_r^{(\tau+1)}$, together with the valid region $\mathbf{n}$, to enhance $\cup_{r=1}^{\tau}\tilde{\mathbf{m}}_r^{(\tau)}$. More precisely, for each $\tau$, the enhancement is carried out in two parts separately: $\cup_{r=1}^{\tau-1}\tilde{\mathbf{m}}_r^{(\tau)}$ is enhanced based on $\cup_{r=1}^{\tau-1}\tilde{\mathbf{m}}_r^{(\tau-1)}$ and $\cup_{r=1}^{\tau-1}\tilde{\mathbf{m}}_r^{(\tau+1)}$ while $\tilde{\mathbf{m}}_{\tau}^{(\tau)}$ is enhanced based on $\tilde{\mathbf{m}}_{\tau}^{(\tau+1)}$. It is also worth mentioning that in our implementation, we decompose
$\mathbf{n}$ into low-level information $\mathbf{l}$ and high-level information $\mathbf{h}$ using a feature extraction netwrok. Let $\hat{\mathbf{m}}_\tau$ denote the enhanced version\footnote{Note that $\hat{\mathbf{m}}_T=\cup_{r=1}^{T}\tilde{\mathbf{m}}_r^{(T)}$ since no enhancement is performed when $\tau=T$.} of $\cup_{r=1}^{\tau}\tilde{\mathbf{m}}_r^{(\tau)}$ $(1\leq\tau\leq T)$. Our algorithm produces $\cup_{\tau=1}^T\hat{\mathbf{m}}_\tau$ as an approximation of $\hat{\mathbf{m}}_{map}$.

\subsection{Feature Extraction Stage}

The feature extraction stage is inspired by Gaussian and Laplacian pyramids. The main component of this stage is a sequence of convolution, Gaussian smoothing, upsampling, another convolution, and subtraction. This sequence will be henceforth referred to as the GLE module. As shown in the first column of Figure \ref{fig:Framework}, the input is first passed through a convolutional layer with 64 kernels, and a ReLU activation function. Next, the generated feature continue passing through a convolutional layer with a number of kernels that is double the number of input channels. Each kernel has a $7\times7$ height and width, $2\times2$ stride, and $3\times3$ padding. It results in a reduced size feature that is then blurred using a $3\times3$ Gaussian kernel moving with a stride of 1 and implementing a padding of 1 to maintain the spatial dimensions fixed. The smoothed feature is passed to the next feature module and to the  upsampling layer, as well. It is upsampled using nearest neighbor to recover the original input size before it is passed through the second convolutional layer. This convolution is characterized with the same hyper-parameters as those of the first one, but it has half the number of kernels recovering the same number of channels as that of the input tensor. The output feature map is produced by subtracting the original input from the feature map coming from the second convolutional layer. Let $I_{\tau-1}$ denotes the input feature maps of ${\tau}_{th}$ GLE module, $\mathcal{G}$ denotes the gaussian smoothing operation, $Up$ denotes the upsample operation, $\mathbf{\Lambda}_{Gs}$ denotes the weights of the convolutional layer before the gaussian smoothing operation, and $\mathbf{\Lambda}_{Up}$ denotes the weights of convolution layer after the upsample operation. The GLE module can be expressed as 
\begin{equation}
	\mathbf{F}_{\tau} = \mathbf{I}_{\tau-1} - \mathbf{\Lambda}_{\mathcal{Q}}(\mathcal{Q}(\mathcal{G}(\mathbf{\Lambda}_{\mathcal{G}}(\mathbf{I}_{\tau-1})))),
\end{equation}
where $\tau \in \{1, 2,...,5\}$. $\mathbf{F}_{\tau}$ denotes output feature maps from ${\tau}_{th}$ GLE module. The $\mathbf{F}_6$ is generated right after the gaussian smooth layer of $5_{th}$ GLE module.

To produce various levels of features, this stage is designed to have 5 GLE modules stacked consecutively, each one feeds into the next. A corrupted image, one with missing pixels and denoted by $\mathbf I_{\text{in}}$ in Figure \ref{fig:Framework}, a corrupted structural image $\mathbf I_{\text{struc}}$ used in \cite{ren2019structureflow} and a binary mask $\mathbf M_{\text{in}}$ are the input to the first module, and the output is the blurred feature volume $\mathbf I_{1}$ as well as the difference feature volume $\mathbf F_1$. $\mathbf I_{1}$ has half the height and width of the input image and double the number of channels, and it is passed to the next GLE module. $\mathbf F_{1}$, on the other hand, is buffered to construct the feature pyramid output that represents the output of the feature extraction stage. The pyramid is formed by stacking the difference-feature volumes generated by each GLE module, namely $\mathbf F_1,\dots, \mathbf F_6$.

\subsection{Iterative Inpainting Stage}
This is the second stage of the proposed solution, which is based on the concept of progressive inpainting. The main elements of this stage are partial convolution, regular convolution, and feature attention. These elements make up two parallel branches, in which features are processed iteratively. The following three subsections detail the inner workings of this stage.

\subsubsection{Partial Convolution}
Partial convolution is a fundamental tool to fill  the irregular holes in deep learning-based image inpainting and keep track of the unfilled regions of the image. To see how a partial convolution layer accomplishes this, let $\mathbf {W}_k\in\mathbb R^{C\times H'\times W'}$ denote the weight tensor of the $k$-th kernel in a partial convolution layer, $\mathbf{X}_{i,j}\in\mathbb R^{C\times H'\times W'}$ denote the input feature patch extracted from the input tensor $\mathbf X_{\text{in}}\in\mathbf R^{C\times H\times W}$ centered around $(i,j)$-th pixels, where $C$, $H'$ and $W'$ are, respectively, the number of channels, height, and width of the patch and $C$, $H$ and $W$ are, respectively, the number of channels, height, and width of the input tensor. Also, let $\widetilde{\mathbf H}_{i,j}$ denote a $H'\times W'$ binary patch centered around the $(i,j)$-th pixel, and $\mathbf H_{i,j}$ is a $C\times H'\times W'$ binary tensor formed by stacking $C$ copies of the matrix $\widetilde{\mathbf H}_{i,j}$. 
Then, the $(i,j)$-th value of the $k$-th output feature map, i.e., $y_{i,j,k}$, produced by a partial convolution layer---before activation---is given by
\begin{equation}
	y_{i,j,k} =
	\begin{cases} g(\mathbf W_k, \mathbf X_{i,j}, \mathbf H_{i,j}) + b, &  \sum_{C,H',W'} \mathbf H_{i,j} > 0,\\
		0, & \text{otherwise},
	\end{cases}
\end{equation}
where $g(\mathbf W_k, \mathbf X_{i,j}, \mathbf H_{i,j})$ is defined as 
\begin{equation}
	\sum_{C,H',W'}\mathbf{W}_k \odot (\mathbf X_{i,j}\odot \mathbf H_{i,j}) \frac{\sum_{C,H',W'}(\mathbf 1)}{\sum_{C,H',W'}\mathbf H_{i,j}},
\end{equation}
$\mathbf 1$ is a $C\times H'\times W'$ tensor of all ones, and $b\in\mathbb R$ is the bias associated with the $k$-th kernel. Following a partial convolution is a mask update to make sure that the mask is keeping up with the updated feature map coming out of the partial convolution. 
Let the full mask be given by
\begin{equation}
	\tilde{\mathbf H} = \left[\begin{array}{ccc}
		h_{11} & \dots & h_{1W} \\
		\vdots & \ddots & \vdots \\
		h_{H1} & \dots & h_{HW}
	\end{array}\right],
\end{equation}
where $H$ and $W$ are, respectively, the height and width of the mask such that $H\gg H'$ and $W\gg W'$, and $\tilde{\mathbf H}
_{i,j}$ is a sub-matrix forming a block in $\tilde{\mathbf H}$ centered around $(i,j)$-th pixel. This mask is updated by convolving an all one kernel with the mask. Let $\mathbf U$ be a $H'\times W'$ kernel of all ones. Then, the updated mask is given by
\begin{equation}
	\tilde{\mathbf H}_{\text{new}} = \tilde{\mathbf H}_{\text{previous}}*\mathbf U,
\end{equation}
where $*$ is the convolution operation with a stride equal to that of the partial convolution kernel. More about partial convolution could be found in \cite{liu2018image}.

\subsubsection{Feature Attention}
For any feature volume $\mathbf F\in\mathbb R^{C\times H\times W}$, an attention tensor could be generated using cosine similarity and softmax. Let $\mathbf f_{i,j}$ and $ \mathbf f_{i^{'},j^{'}}$ denote pair of feature values at location $i,j$ and $i^{'}$,$j^{'}$. Then, their cosine similarity is computed as follows:
\begin{equation}
	z_{i,j,i^{'},j^{'}} = \langle \frac{\mathbf f_{i,j}}{\|\mathbf f_{i,j}\|}, \frac{\mathbf f_{i^{'},j^{'}}}{\|\mathbf f_{i^{'},j^{'}}\|}\rangle,
\end{equation}
where $z_{i,j,i^{'},j^{'}}$ denotes the cosine similarity score between the $\mathbf f_{i,j}$ and $\mathbf f_{i^{'},j^{'}}$ \footnote{ $\mathbf f_{i,j}$ is a feature vector with size $1\times1\times C$ at position $i,j$ from a feature volume with size $H\times W \times C$. $\mathbf f_{i,j}$ and $\mathbf f_{i^{'},j^{'}}$ have the same size, and $\|\mathbf f_{i,j}\|$ and $\|\mathbf f_{i^{'},j^{'}}\|$ are the second norms of those vectors.}. Let $\mathbf Z_{i,j}\in\mathbb R^{H\times W}$ denotes the score matrix of a feature vector at location $i,j$ and all $C$-dimensional feature vectors in $\mathbf F$. The softmax function is applied across the height and width to generate the attention score of location $i,j$ in $\mathbf F$. Formally, this is expressed as follows: 
\begin{equation}
	\hat{\mathbf Z}_{i,j} = \mathrm{sfm}(\mathbf Z_{i,j}),
\end{equation}
where $\hat{\mathbf Z}_{i,j}\in\mathbb R^{H\times W}$. The final feature volume $\mathbf F$ has an attention tensor $\hat{\mathbf Z}\in\mathbb R^{HW\times H\times W}$ formed by stacking $HW$ score maps $\hat{\mathbf Z}_{i,j}$. Based on the calculated score map, we reuse the feature patches from the input of feature attention module as de-convolutional filters to reconstruct the new feature map.

\subsubsection{Putting It All Together}
Iterative inpainting is built on top of the feature extraction stage with the feature pyramid as its input. This is illustrated in the middle column of Figure \ref{fig:Framework}. The pyramid is first split into two halves; feature maps coming from the first three feature modules (i.e., the first three from the input side) are concatenated to form the feature volume $\mathbf F_{\text{low}}\in\mathbb R^{C_{in}\times H_{in}\times W_{in}}$ with $C_{in}$ channels, $H_{in}$ height, and $W_{in}$ width. Feature maps coming from the last three feature modules form another feature volume denoted $\mathbf F_{\text{high}}\in\mathbb R^{C_{in}\times H_{in}\times W_{in}}$. Those two volumes are sent down two different but parallel iterative branches that have the same composition of layers. Both start with two partial convolutions with leaky ReLU activations, followed by a feature attention module. The specifications of each layer are detailed at the bottom of the middle column of Figure \ref{fig:Framework}.

Each branch processes the input volume iteratively, which is done as follows. Let $\tau$ represent a time index for the iterative process.
Both $\mathbf F_{\text{low}}(0)$ or $\mathbf F_{\text{high}}(0)$ goes through the partial convolutions and the attention module, making up the first iteration ($\tau=1$). The outputs, denoted $\mathbf F_{\text{low}}(\tau+1)$ and $\mathbf F_{\text{high}}(\tau+1)$, are used to initialize the next iteration as well as construct a new feature volume. A copy of $\mathbf F_{\text{low}}(\tau+1)$ and $\mathbf F_{\text{high}}(\tau+1)$ is sent back to the input to undergo the next iteration. Another copy is sent forward to a concatenation operation to form part of a new feature volume denoted $\mathbf F_{\text{cat}}$. This keeps on going for $T$ iterations ($\tau\in\{1,2,\dots,T\}$) until $\mathbf F_{\text{cat}}$ is complete, i.e., a tensor of dimensions $C_{\text{cat}}\times H_{\text{cat}}\times W_{\text{cat}}$ where $C_{\text{cat}} = 2TC_{in}$ is formed. This tensor is, finally, passed to a convolution layer with leaky ReLU activation, which generates the intermediate feature volume $\mathbf F_{\text{int}}$.

\textbf{Remark:} Please note that the $\mathbf F_{\text{int}}$ feature volume comprises $C_{\text{int}} = C_{\text{cat}} = 2TC_{in}$ feature maps, which could be split into $T$ sub-volumes. This is important for the sake of the third and final stage of the proposed architecture.

\subsection{Enhancement and Reconstruction Stage}\label{sec:enh}
\subsubsection{Reinpainting Component}
\label{section:Re}
The main idea behind the reinpainting component is to re-enhance the fused feature sub-volumes in $\mathbf{F}_{\text{int}}$. This is done along two branches that process two different concatenations of feature sub-volumes from $\mathbf{F}_{\text{int}}$. See Figure \ref{fig:Framework}. 
Let $\mathbf{F}_{\text{int}}(\tau)$ represent the $\tau$-th sub-volume in $\mathbf{F}_{\text{int}}$, where $\tau\in\{1,\dots,T-1\}$. The first branch concatenates $\mathbf{F}_{\text{int}}(\tau-1)$, $\mathbf{F}_{\text{int}}(\tau)$, and $\mathbf{F}_{\text{int}}(\tau+1)$ and passes them into three convolutional layers with ReLU activations. The result is multiplied with the updated mask of iteration $\tau-1$ from the second stage, i.e., $\mathbf{\tilde H} (\tau-1)$, to eliminate the negative effect of the unfilled region in each iteration.

The second branch is symmetric with the first, but focuses on different sub-volumes. It concatenates $\mathbf{F}_{\text{int}}(\tau)$ and $\mathbf{F}_{\text{int}}(\tau+1)$ and passes them through three convolutional layers with ReLU activations. The result here is multiplied with the difference of two updated masks from iterations $\tau$ and $\tau-1$, i.e., $\mathbf{\tilde H} (\tau) - \mathbf{\tilde H} (\tau-1)$, which only contains information from the intersection region between $\mathbf{F}_{\text{int}}(\tau)$ and $\mathbf{F}_{\text{int}}(\tau-1)$. The results of the two branches are combined with the sub-volume $\mathbf{F}_{\text{int}}(\tau)$ to produce a new sub-volume $\mathbf{F}_{\text{reinp}}(\tau)$.

\subsubsection{Reconstruction Component}
The reinpainting model outputs a feature volume $\mathbf F_{\text{reinp}}$ that is fed to the reconstruction component. This is the final component of the proposed architecture, responsible for producing the complete image. A visualization of the reconstruction component is illustrated in the left panel of Fig. \ref{fig:Framework}. This component adopts the feature merging module from \cite{li2020recurrent} which fuses the feature group based on the filled locations in each iteration. The merge module feeds into three upsampling layers followed by a partial convolution layer, three residual blocks, and a sequence of three convolutional layers. The complete architecture is summarized in Algorithm \ref{alg:Proposed} \footnote{To simplify the notation in Algorithm \ref{alg:Proposed}, We use $\mathcal{I}$ to express Iterative Inpainting Module, use $\mathcal{RI}$ to express Re-Inpainting Module}.
\begin{algorithm}[h]
	\caption{The Proposed Inpainting Network} 
	\label{alg:Proposed}
	\hspace*{\algorithmicindent} \textbf{Input} Input image $I_{in}$, \\
	\hspace*{1.43cm}Input structural $I_{struc}$, Input mask $M_{in}$,\\ 
	\hspace*{1.43cm}Total number of iteration $T$\\
	\hspace*{\algorithmicindent} \textbf{Output} Recovered image $I_{out}$\\
	\begin{algorithmic}[1]
		\vspace*{-.4cm}
		\STATE $\mathbf F_{\text{high}}(0), \mathbf F_{\text{low}}(0), \mathbf{\tilde H} (0) \leftarrow GLE(I_{in}, I_{struc}, M_{in})$
		\STATE $FeaturePool \leftarrow \{\mathbf F_{\text{high}}(0), \mathbf F_{\text{low}}(0)\}$
		\STATE $\tau \leftarrow0$
		\IF{$\tau$ $\leq$ $T$} 
		\STATE $\mathbf F_{\text{low}}(\tau+1), \mathbf{\tilde H} (\tau+1) \leftarrow \mathcal{I}(\mathbf F_{\text{low}}(\tau),\mathbf{\tilde H} (\tau))$
		\STATE $\mathbf F_{\text{high}}(\tau+1), \mathbf{\tilde H} (\tau+1) \leftarrow \mathcal{I}(\mathbf F_{\text{high}}(\tau),\mathbf{\tilde H} (\tau))$
		\STATE $\mathbf{F}_{\text{int}}(\tau+1) \leftarrow FeatureFuse(\mathbf F_{\text{high}}(\tau+1),\mathbf F_{\text{low}}(\tau+1))$
		\STATE $\tau \gets \tau+1$ 
		\ENDIF
		\STATE $\tau \leftarrow 1$
		\IF{$\tau$ $\leq$ $T-1$}
		\STATE $F = \{\mathbf{F}_{\text{int}}(\tau-1), \mathbf{F}_{\text{int}}(\tau), \mathbf{F}_{\text{int}}(\tau+1)$ 
		\STATE $H = \{\mathbf{\tilde H} (\tau-1), \mathbf{\tilde H} (\tau), \mathbf{\tilde H} (\tau+1)\}$
		\STATE $\mathbf{F}_{\text{reinp}}(\tau) \leftarrow \mathcal{RI}(F,H) + \mathbf{F}_{\text{int}}(\tau)$
		\STATE $FeaturePool \leftarrow FeaturePool + \{\mathbf{F}_{\text{reinp}}(\tau)\}$
		\STATE $\tau \gets \tau+1$
		\ENDIF
		\STATE $F_{merged} \leftarrow FeatureMerge(FeaturePool)$
		\STATE $I_{out} \leftarrow Reconstruction(F_{merged})$
		\RETURN $I_{out}$
	\end{algorithmic}
\end{algorithm}

\subsection{Loss Functions}
This section describes the loss functions for training the proposed inpainting network. It is a composite loss with multiple terms accounting for different aspects that the proposed algorithm needs to maintain. Perceptual loss and style loss are two of those terms that are popular for solving image generation problems. They are calculated using groundtruth and output feature maps obtained from a pretrained VGG model \cite{simonyan2014very}. \textit{Groundtruth features} are those produced by the max-pooling layers of the VGG network when the input is the complete groundtruth image whereas \textit{output features} are those obtained from the same pooling layers but with the restored image as an input. Formally, the perceptual loss is given by

\begin{equation}
	L_{\text{perc}} = \sum_{\theta=1} ^{N} \frac{1}{H_{\theta}W_{\theta}C_{\theta}}|\phi^{gt}_{\theta} - \phi^{out}_{\theta}|_1,
\end{equation}
and the style loss is given by
\begin{equation}
	\begin{split}
		L_{\text{sty}} = \sum_{\theta=1} ^{N} \frac{1}{C_{\theta} \times C_{\theta}} \bigg|\frac{1}{H_{\theta}W_{\theta}C_{\theta}}(\phi^{gt}_{\theta} (\phi^{gt}_{\theta})^{T} \\- \phi^{out}_{\theta}(\phi^{out}_{\theta})^T)\bigg|_1,
	\end{split}
\end{equation}
where $\phi^{gt}_{\theta}$ denotes the vectorized groundtruth feature map from the $\theta_{th}$ pooling layer of VGG-16, $\phi^{out}_{\theta}$ denotes the vectorized output feature map from the $\theta_{th}$ pooling layer of pretrained VGG-16, and $C_\theta$, $H_\theta$, and $W_\theta$ are, respectively, the number of channels, height, and width of the $\theta_{th}$ feature map.

The third term of the composite loss is the total variation loss, which enforces smoothness in the region of predicted pixels (i.e., the holes) \cite{mahendran2015understanding,liu2018image}. Formally, this term is formulated as follows. Let $\mathbf I_{\text{out}}^{i,j}$ denote the pixel value of output image at location $i$, $j$, $N$ denote the total number of elements in the output image, $\mathcal R$ denote the set of pixels surrounding a corrupted pixel $\mathbf I_{\text{out}}^{i,j}$. The total variation loss is given by
\begin{equation}
	\begin{split}
		L_{tv} = \sum_{(i,j)\in R , (i,j+1) \in R} \frac{|I^{i,j+1} - I^{i,j}|_{1}}{N} \\ 
		+ \sum_{(i,j)\in R , (i+1,j) \in R} \frac{|I^{i+1,j} - I^{i,j}|_{1}}{N}.
	\end{split}
\end{equation}

The last two terms in the composite loss are first norms of the difference between the output and groundtruth images. Let $I_{out}$ denote the output image from the proposed algorithm, $I_{gt}$ denote the ground truth image, and $\mathbf{\tilde{H}}_{\text{gt}}$ denote the groundtruth mask of the image. The two terms are, then, given by 
\begin{equation}
	L_{\text{valid}} = |I_{out}\odot \mathbf{\tilde{H}}_{\text{gt}}-I_{gt}\odot \mathbf{\tilde{H}}_{\text{gt}}|_{1},
\end{equation}
\begin{equation}
	L_{\text{hole}} = |I_{out}\odot (1-\mathbf{\tilde{H}}_{\text{gt}}) -I_{gt}\odot (1-\mathbf{\tilde{H}}_{\text{gt}})|_{1},
\end{equation}
where $L_{\text{valid}}$ expresses the first norm loss between undamaged region of the output image and the ground truth image, and $L_{\text{hole}}$ expresses the first norm loss between filled region of the output image and the ground truth image. The composite loss, as the name suggests, is a weighted sum of all the above terms \begin{equation}
	\begin{split}
		\mathcal L = \lambda_{\text{valid}} L_{\text{valid}} + \lambda_{\text{hole}} L_{\text{hole}} + \lambda_{\text{perc}} L_{\text{perc}} \\
		+ \lambda_{\text{style}} L_{\text{style}} + \lambda_{tv} L_{tv},
	\end{split}
\end{equation}
where $\lambda_{\text{valid}}, \lambda_{\text{hole}}, \lambda_{\text{perc}}, \lambda_{\text{style}}$, and $\lambda_{tv}$ are all hyper-parameters scaling the contribution of each of their respective terms to the composite loss.

\begin{figure}[htp] 
	\centering
	\stackunder[1pt]{\includegraphics[width=0.5\textwidth]{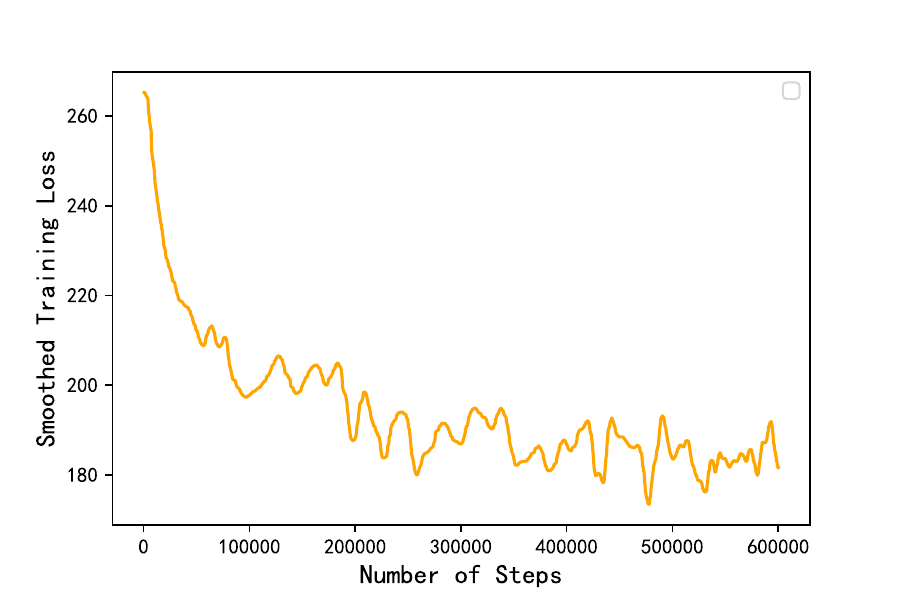}}{}%
	\hfill%
	\hspace{-6mm}
	\stackunder[1pt]{\includegraphics[width=0.5\textwidth]{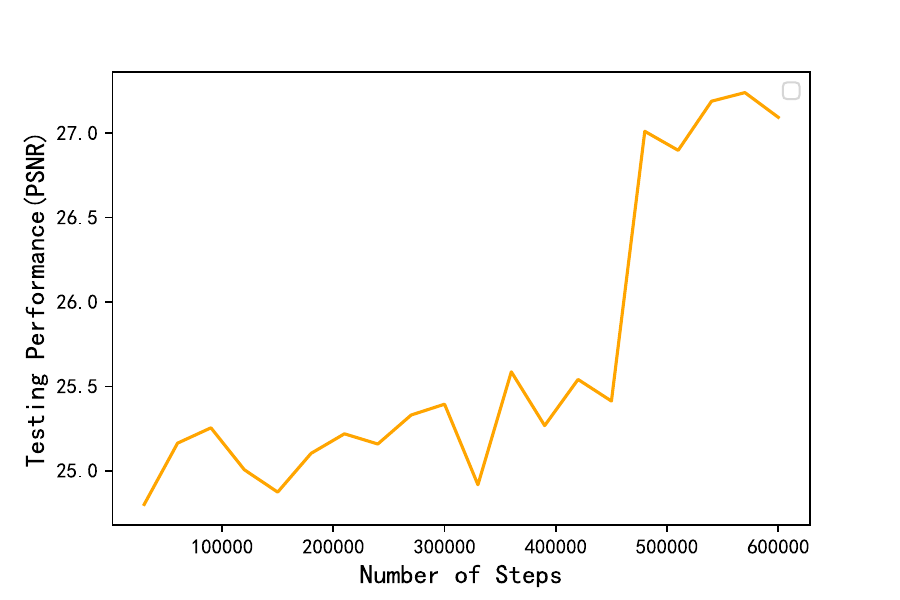}}{}%
	\caption{The history graph of Smoothed training loss and testing performance vs.\ number of steps on Celeba dataset.}
	\label{fig:trainingpic}
\end{figure}

\section{Experimental Setup}\label{sec:exp_setup}
The proposed algorithm needs to be put to test in order to demonstrate its performance. This section presents the experimental setup adopted to evaluate its performance. It describes the development datasets, the implementation details, and the benchmark algorithms.

\begin{table*}[!htbp]
	\centering
	\begin{threeparttable}
		\centering
		\caption{Numerical comparisons on three datasets.}
		\begin{tabular}{c|c|c|c|c|c|c|c|c|c|c|c|c|c}
			\hline
			\multicolumn{2}{c|}{} & \multicolumn{4}{c|}{PSNR} & \multicolumn{4}{c|}{SSIM} & \multicolumn{4}{c}{Mean $l_1$}\\
			\cline{1-14}
			\multicolumn{2}{c|}{Mask Ratio} & 10-20\%  & 30-40\%  & 40-50\% & 50-60\%  & 10-20\%  & 30-40\%  & 40-50\% & 50-60\%  & 10-20\%  & 30-40\%  & 40-50\%  & 50-60\% \\
			\cline{1-14}
			\multirow{8}{*}{PVS} & {PIC\cite{zheng2019pluralistic}} & {29.02} & {24.82} & {23.71} & {19.61} & {0.930} & {0.795} & {0.757} & {0.524} & {0.0134} & {0.0327} & {0.0429} &  {0.0785}\\
			{} & {PC\cite{liu2018image}} & {30.42} & {25.35} & {23.86} & {21.47} & {0.933} & {0.825} & {0.763} & {0.628} & {0.1001} & {0.0253} & {0.0373} & {0.0612} \\
			{} & {EC\cite{nazeri2019edgeconnect}} & {30.83} & {25.94} & {24.03} & {21.76} & {0.945} & {0.849} & {0.781} & {0.653}  & {0.0084} & {0.0220} & {0.0297} & {0.0578} \\
			{} & {PRVS\cite{li2019progressive}} & {30.87} & {26.21} & {24.11} & {21.94} & {0.952} & {0.857} & {0.794} & {0.665} & {0.0085} & {0.0217} & {0.0288} & {0.0567}\\
			{} & {RFR\cite{li2020recurrent}} & {31.74} & {26.35} & {24.56} & {22.43} & {0.954} & {0.864} & {0.811} & {0.687} & {0.0076} & {0.0211} & {0.0280} & {0.0543}\\
			{} & {MDEFE\cite{liu2020rethinking}} & {31.11} & {26.23} & {24.18} & {22.17} & {0.954} & {0.859} & {0.806} & {0.674} & {0.0081} & {0.0217} & {0.0287} & {0.0549} \\
			{} & {Ours} & \textbf{31.99} & \textbf{26.86} & \textbf{25.04} & \textbf{22.61} & \textbf{0.964} & \textbf{0.889} & \textbf{0.833} & \textbf{0.721} &  \textbf{0.0073} & \textbf{0.0192} & \textbf{0.0268} & \textbf{0.0485} \\
			\cline{1-14}
			\multirow{8}{*}{Celeba} & {PIC\cite{zheng2019pluralistic}} & {30.69} & {24.65} & {21.40} & {19.16} & {0.965} & {0.875} & {0.739} & {0.667} & {0.0094} & {0.0258} & {0.0433} & {0.0756} \\
			{} & {PC\cite{liu2018image}} & {32.72} & {26.59} & {24.24} & {22.03} & {0.970} & {0.915} & {0.864} &  {0.782} & {0.0062} & {0.0181} & {0.0269} & {0.0537}\\
			{} & {EC\cite{nazeri2019edgeconnect}} & {32.47} & {26.61} & {24.43} & {21.42} & {0.974} & {0.921} & {0.873} & {0.746} & {0.0065} & {0.0185} & {0.0260} & {0.0564} \\
			{} & {PRVS\cite{li2019progressive}} & {33.12} & {27.05} & {24.82} & {22.32} & {0.977} & {0.928} & {0.883} & {0.793} & {0.0059} & {0.0169} & {0.0252} & {0.0493} \\
			{} & {RFR\cite{li2020recurrent}} & {33.58} & {27.62} & {25.48} & {22.59} & {0.980} & {0.931} & {0.896} & {0.822} & {0.0057} & {0.0161} & {0.0224} & {0.0482}\\
			{} & {MDEFE\cite{liu2020rethinking}} & {33.20} & {27.21} & {24.97} & {22.44} & {0.976} & {0.925} & {0.879} & {0.814} & {0.0059} & {0.0167} & {0.0251} & {0.0509}\\
			{} & {Ours} & \textbf{33.82} & \textbf{27.93} & \textbf{25.92} & \textbf{22.76} & \textbf{0.984} & \textbf{0.945} & \textbf{0.915} & \textbf{0.834} &  \textbf{0.0053} & \textbf{0.0150} & \textbf{0.0213} &  \textbf{0.0359}\\
			\cline{1-14}
			\multirow{8}{*}{Place2} & {PIC\cite{zheng2019pluralistic}}  & {26.67} & {21.24} & {19.13} & {17.27} & {0.931} & {0.761} & {0.656} & {0.499} & {0.0137} & {0.0342} & {0.0573} & {0.0953}\\
			{} & {PC\cite{liu2018image}} & {26.84} & {22.25} & {20.09} & {18.33} & {0.935} & {0.768} & {0.722} & {0.538} & {0.0131} & {0.0312} & {0.0475} & {0.0861} \\
			{} & {EC\cite{nazeri2019edgeconnect}} & {27.02} & {22.41} & {20.35} & {18.41} & {0.934} & {0.811} & {0.745} & {0.531} & {0.0135} & {0.0309} & {0.0447} & {0.0846}  \\
			{} & {PRVS\cite{li2019progressive}} & {27.35} & {22.73} & {20.51} & {18.72} & {0.937} & {0.825} & {0.773} & {0.569} & {0.0140} & {0.0293} & {0.0423} & {0.0769} \\       
			{} & {RFR\cite{li2020recurrent}} & {27.90} & {23.27} & {21.34} & {18.87} & {0.945} & {0.837} & {0.782} & {0.589}  & {0.0114} & {0.0284} & {0.0402} & {0.0726} \\
			{} & {MDEFE\cite{liu2020rethinking}} & {27.55} & {22.92} & {21.01} & {18.65} & {0.941} & {0.832} & {0.775} & {0.582} & {0.0126} & {0.0297} & {0.0411} & {0.0737}  \\
			{} & {Ours} & \textbf{28.89} & \textbf{23.66} & \textbf{21.94} & \textbf{19.59} & \textbf{0.957} & \textbf{0.864} & \textbf{0.800} & \textbf{0.667} & \textbf{0.0099} & \textbf{0.0266} & \textbf{0.0368} & \textbf{0.0552}\\
			\cline{1-14}
		\end{tabular}
		\label{Tab:T_1}
		\begin{tablenotes}
			\item[1] The best performances are highlighted in \textbf{bold}.
		\end{tablenotes}
	\end{threeparttable}
\end{table*}

\subsection{Datasets}
Four development datasets are adopted here:
\begin{itemize}
	\item \textbf{Paris Streetview Dataset}\cite{doersch2012makes} is collected from Google StreetView, a large-scale dataset that includes street images for 12 cities across the world. This dataset contains 15000 images, 14900 images for training, and 100 for testing.
	\item \textbf{Large-scale CelebFaces Attributes (CelebA) Dataset}\cite{liu2015deep} is a well-known and publicly available face recognition dataset. It includes around 200K celebrity images representing 10000 different identities, all of which have a wide range of posture variations. This dataset contains 202599 images, 162770 images for training, 19867 images for validation, and 19962 images for testing.
	\item \textbf{Place2 Dataset}\cite{liu2015deep} This dataset contains 8 million images which are collected from 365 scene categories, like streets, indoor rooms and so on. 
	\item \textbf{NVIDIA Irregular Mask Dataset Dataset}\cite{nv_irregular_maskdata} is a popular irregular mask dataset. This dataset contains 12000 irregular masks which are randomly drawn by individuals. The mask ratio of the dataset is uniformly distributed on 0.0$\sim$0.1, 0.1$\sim$0.2, 0.2$\sim$0.3, 0.3$\sim$0.4, 0.4$\sim$0.5, and 0.5$\sim$0.6. Each mask ratio class has 1000 masks with and without border.
\end{itemize}

\subsection{Implementation Details}
The proposed algorithm is trained with batch size of 4 on two NVIDIA 1080 TITANs. We use corrupted images, structural maps, and irregular holes as inputs, which are resized to 256 $\times$ 256. Adam \cite{Diederik2015Adam} optimizer is used to train the network. The training is conducted with a learning rate of $10\time 10^{-4}$, and the network is fine-tuned with a learning rate of $10\time 10^{-5}$. The network is trained on Paris and CelebA Dataset for 40 epochs and fine-tuned for 20 epochs. On Place2 Dataset, the network is trained for 200 epochs and fine-tuned for 100 epochs. During the fine-tuning, only the weights of batch normalization layers are frozen while the rest are adjusted. The hyper-parameters of the loss function are set to $\lambda_{\text{valid}} = 1$, $\lambda_{\text{hole}} = 6$, $\lambda_{\text{perc}} = 0.05$, $\lambda_{\text{style}} = 120$, and $\lambda_{tv} = 0.1$. History graphs of smoothed training loss and testing performance vs.\ number of steps on Celeba dataset are shown in Fig.\ref{fig:trainingpic}. The total number of parameters of the proposed network is 82 million. This means its memory footprint assuming float-32 representation is roughly 312 MB and its inference time averages 0.147 ms per image.

\begin{figure*}[!htbp] 
	\begin{minipage}[b]{\linewidth} 
		\subfloat[Input]{
			\begin{minipage}[b]{0.111\linewidth} 
				\includegraphics[width=\linewidth]{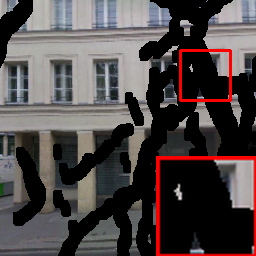}\vspace{4pt}
				\includegraphics[width=\linewidth]{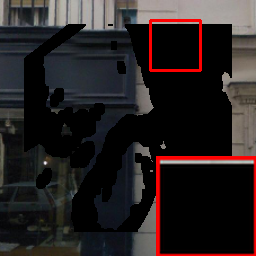}
			\end{minipage}
		}
		\hfill
		\subfloat[PIC\cite{zheng2019pluralistic}]{
			\begin{minipage}[b]{0.111\linewidth}
				\includegraphics[width=\linewidth]{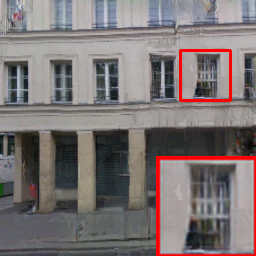}\vspace{4pt}
				\includegraphics[width=\linewidth]{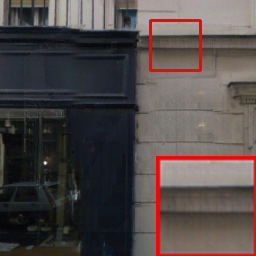}
			\end{minipage}
		}
		\hfill
		\subfloat[PC\cite{liu2018image}]{
			\begin{minipage}[b]{0.111\linewidth}
				\includegraphics[width=\linewidth]{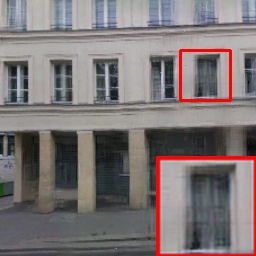}\vspace{4pt}
				\includegraphics[width=\linewidth]{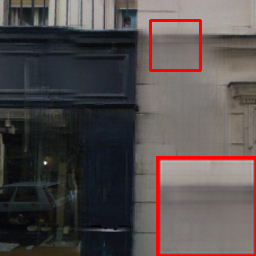}
			\end{minipage}
		}
		\hfill
		\subfloat[EC\cite{nazeri2019edgeconnect}]{
			\begin{minipage}[b]{0.111\linewidth}
				\includegraphics[width=\linewidth]{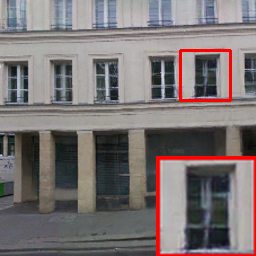}\vspace{4pt}
				\includegraphics[width=\linewidth]{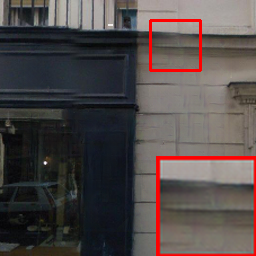}
			\end{minipage}
		}
		\hfill
		\subfloat[PRVS\cite{li2019progressive}]{
			\begin{minipage}[b]{0.111\linewidth}
				\includegraphics[width=\linewidth]{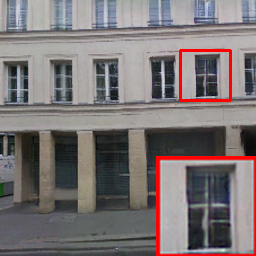}\vspace{4pt}
				\includegraphics[width=\linewidth]{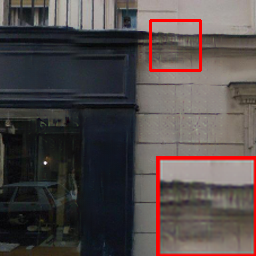}
			\end{minipage}
		}
		\hfill
		\subfloat[RFR\cite{li2020recurrent}]{
			\begin{minipage}[b]{0.111\linewidth}
				\includegraphics[width=\linewidth]{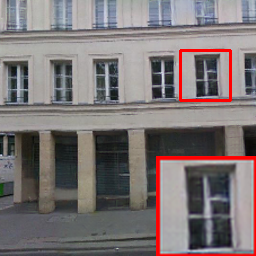}\vspace{4pt}
				\includegraphics[width=\linewidth]{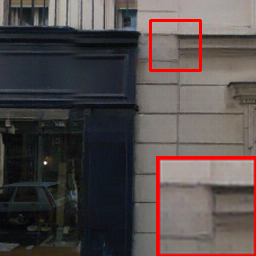}
			\end{minipage}
		}
		\hfill
		\subfloat[Ours]{
			\begin{minipage}[b]{0.111\linewidth}
				\includegraphics[width=\linewidth]{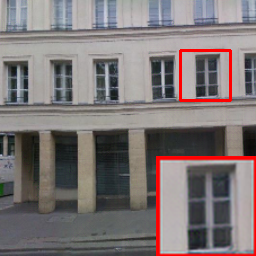}\vspace{4pt}
				\includegraphics[width=\linewidth]{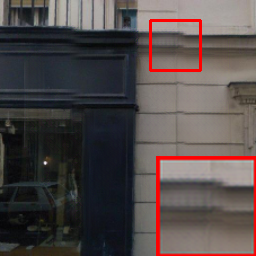}
			\end{minipage}
		}
		\hfill
		\subfloat[GT]{
			\begin{minipage}[b]{0.111\linewidth}
				\includegraphics[width=\linewidth]{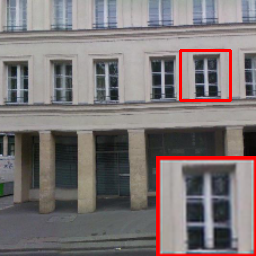}\vspace{4pt}
				\includegraphics[width=\linewidth]{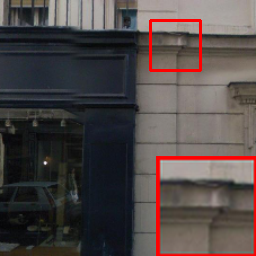}
			\end{minipage}
		}
	\end{minipage}
	\vfill
	\caption{Visual results on Paris StreetView.}
	\label{fig:Paris}
\end{figure*}
\begin{figure*}[!htbp] 
	\begin{minipage}[b]{\linewidth} 
		\subfloat[Input]{
			\begin{minipage}[b]{0.111\linewidth} 
				\includegraphics[width=\linewidth]{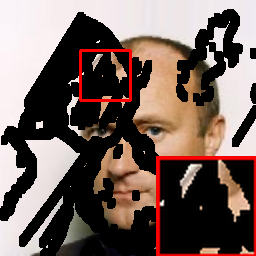}\vspace{4pt}
				\includegraphics[width=\linewidth]{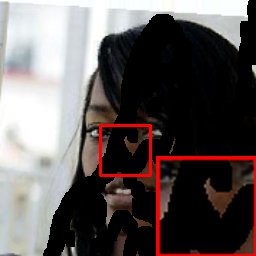}
			\end{minipage}
		}
		\hfill
		\subfloat[PIC\cite{zheng2019pluralistic}]{
			\begin{minipage}[b]{0.111\linewidth}
				\includegraphics[width=\linewidth]{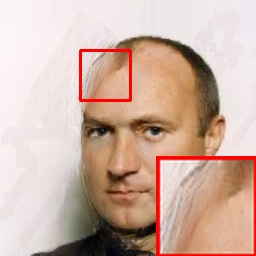}\vspace{4pt}
				\includegraphics[width=\linewidth]{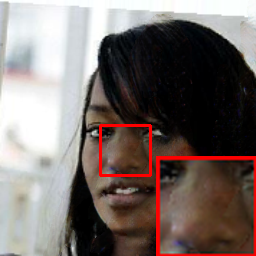}
			\end{minipage}
		}
		\hfill
		\subfloat[PC\cite{liu2018image}]{
			\begin{minipage}[b]{0.111\linewidth}
				\includegraphics[width=\linewidth]{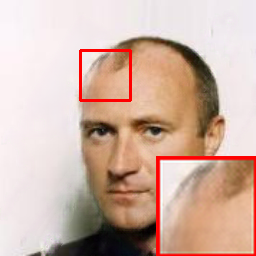}\vspace{4pt}
				\includegraphics[width=\linewidth]{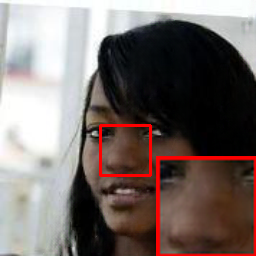}
			\end{minipage}
		}
		\hfill
		\subfloat[EC\cite{nazeri2019edgeconnect}]{
			\begin{minipage}[b]{0.111\linewidth}
				\includegraphics[width=\linewidth]{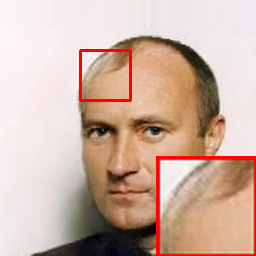}\vspace{4pt}
				\includegraphics[width=\linewidth]{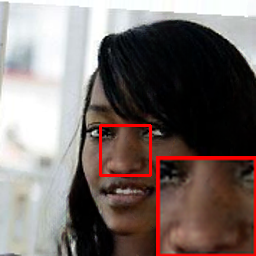}
			\end{minipage}
		}
		\hfill
		\subfloat[PRVS\cite{li2019progressive}]{
			\begin{minipage}[b]{0.111\linewidth}
				\includegraphics[width=\linewidth]{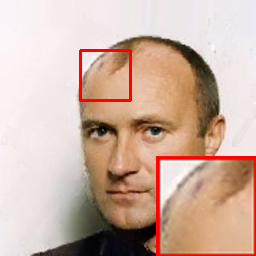}\vspace{4pt}
				\includegraphics[width=\linewidth]{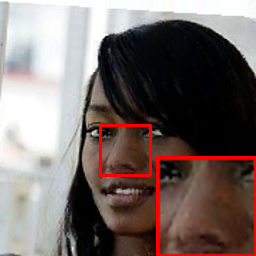}
			\end{minipage}
		}
		\hfill
		\subfloat[RFR\cite{li2020recurrent}]{
			\begin{minipage}[b]{0.111\linewidth}
				\includegraphics[width=\linewidth]{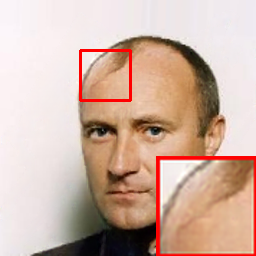}\vspace{4pt}
				\includegraphics[width=\linewidth]{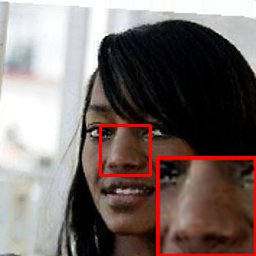}
			\end{minipage}
		}
		\hfill
		\subfloat[Ours]{
			\begin{minipage}[b]{0.111\linewidth}
				\includegraphics[width=\linewidth]{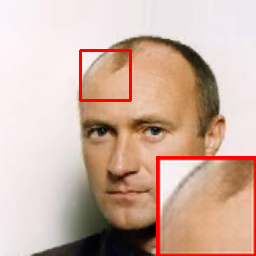}\vspace{4pt}
				\includegraphics[width=\linewidth]{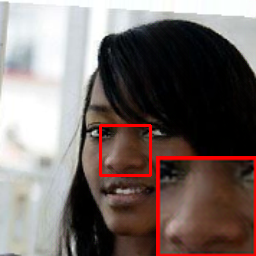}
			\end{minipage}
		}
		\hfill
		\subfloat[GT]{
			\begin{minipage}[b]{0.111\linewidth}
				\includegraphics[width=\linewidth]{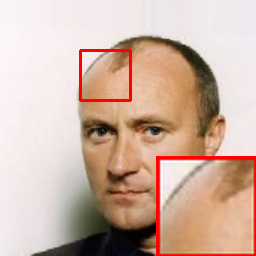}\vspace{4pt}
				\includegraphics[width=\linewidth]{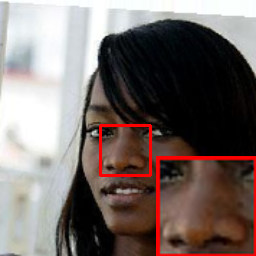}
			\end{minipage}
		}
	\end{minipage}
	\vfill
	\caption{Visual results on CelebA.}
	\label{fig:CelebA}
\end{figure*}
\begin{figure*}[!htbp] 
	\begin{minipage}[b]{\linewidth} 
		\subfloat[Input]{
			\begin{minipage}[b]{0.111\linewidth} 
				\includegraphics[width=\linewidth]{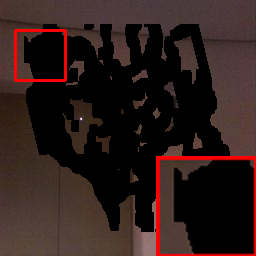}\vspace{4pt}
				\includegraphics[width=\linewidth]{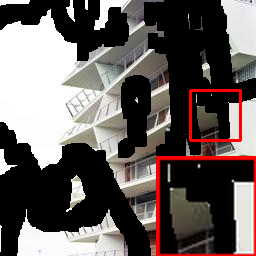}
			\end{minipage}
		}
		\hfill
		\subfloat[PIC\cite{zheng2019pluralistic}]{
			\begin{minipage}[b]{0.111\linewidth}
				\includegraphics[width=\linewidth]{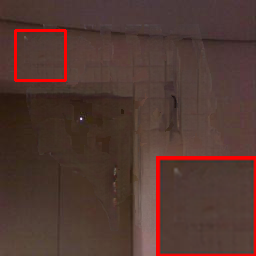}\vspace{4pt}
				\includegraphics[width=\linewidth]{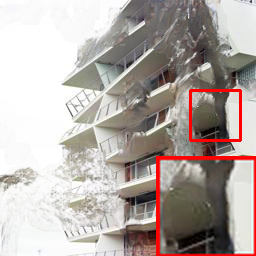}
			\end{minipage}
		}
		\hfill
		\subfloat[PC\cite{liu2018image}]{
			\begin{minipage}[b]{0.111\linewidth}
				\includegraphics[width=\linewidth]{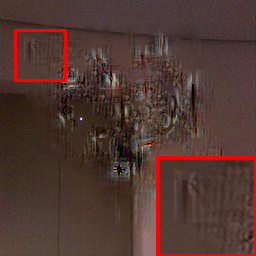}\vspace{4pt}
				\includegraphics[width=\linewidth]{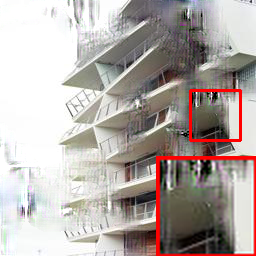}
			\end{minipage}
		}
		\hfill
		\subfloat[EC\cite{nazeri2019edgeconnect}]{
			\begin{minipage}[b]{0.111\linewidth}
				\includegraphics[width=\linewidth]{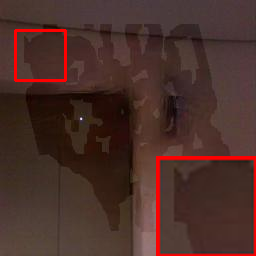}\vspace{4pt}
				\includegraphics[width=\linewidth]{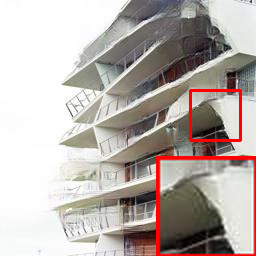}
			\end{minipage}
		}
		\hfill
		\subfloat[PRVS\cite{li2019progressive}]{
			\begin{minipage}[b]{0.111\linewidth}
				\includegraphics[width=\linewidth]{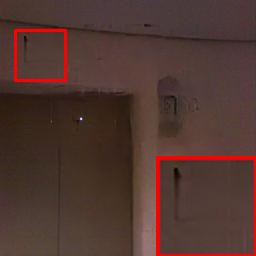}\vspace{4pt}
				\includegraphics[width=\linewidth]{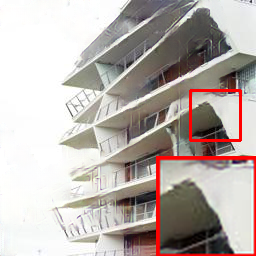}
			\end{minipage}
		}
		\hfill
		\subfloat[RFR\cite{li2020recurrent}]{
			\begin{minipage}[b]{0.111\linewidth}
				\includegraphics[width=\linewidth]{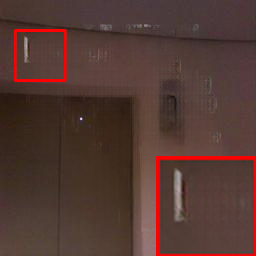}\vspace{4pt}
				\includegraphics[width=\linewidth]{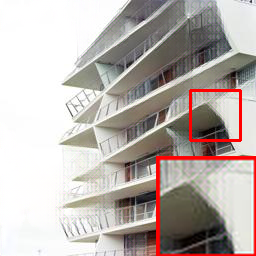}
			\end{minipage}
		}
		\hfill
		\subfloat[Ours]{
			\begin{minipage}[b]{0.111\linewidth}
				\includegraphics[width=\linewidth]{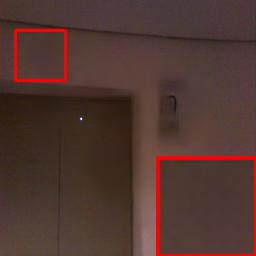}\vspace{4pt}
				\includegraphics[width=\linewidth]{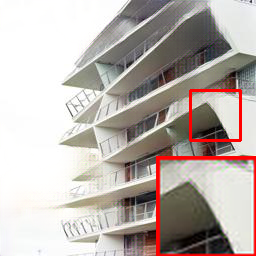}
			\end{minipage}
		}
		\hfill
		\subfloat[GT]{
			\begin{minipage}[b]{0.111\linewidth}
				\includegraphics[width=\linewidth]{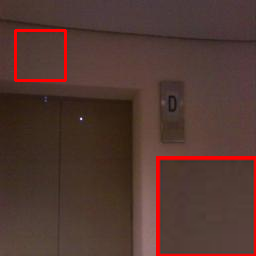}\vspace{4pt}
				\includegraphics[width=\linewidth]{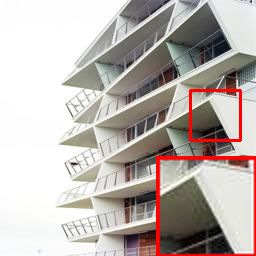}
			\end{minipage}
		}
	\end{minipage}
	\vfill
	\caption{Visual results on Place2.}
	\label{fig:Place2}
\end{figure*}

\subsection{Benchmark Algorithms and Evaluation Metrics}
The proposed algorithm is compared to five state-of-the-art methods, namely  PIC\cite{zheng2019pluralistic}, PC\cite{liu2018image}, PRVS\cite{li2019progressive}, EC\cite{nazeri2019edgeconnect}, RFR\cite{li2020recurrent} and MDEFE \cite{liu2020rethinking}. We use PIC as baseline which is a probabilistically principled framework in image inpainting. PC is a fundamental technique that can be considered as another baseline in image inpainting. EC is a two-stage image inpainting method based on the edge recovering method. PRVS and RFR belong to the family of progressively image inpainting. The PRVS progressively recover the image structural information while RFR recover the image contextual information.  Those five algorithms are henceforth referred to as the benchmark algorithms. The proposed algorithm is compared to all five using three metrics, which are Peak Signal-to-Noise Ratio (PSNR), Structural Similarity Index (SSIM) and mean first norm loss ($L_1$).

\section{Evaluation Results}
The performance of the proposed algorithm is evaluated in this section using the setup described in Section \ref{sec:exp_setup}. The evaluation starts with quantitative analysis where the proposed algorithm is benchmarked to others. Then, a qualitative analysis follows. It presents a comparison of the quality of inpainted images between the proposed algorithm and the benchmark algorithms. Finally, this section is concluded with an ablation analysis illustrating the value of each novel component in the proposed network.

\subsection{Quantitative Analysis}
The proposed algorithm is compared to the benchmark algorithms on the basis of PSNR, SSIM, and $L_1$ loss. Table~\ref{Tab:T_1} presents the comparison results on the Paris StreetView, CelebA, and Place2 datasets. It presents the results for different choices of masking percentage (i.e., mask ratio). The performance of the proposed algorithm stands out throughout the table; despite the slim margin in some cases, its performance could be argued to best all other competing algorithms on all three datasets.

\subsection{Qualitative Analysis}
The above quantitative results are translated into visual analysis to demonstrate the inpainting quality of the proposed algorithm. This is done through a few examples from three datasets, namely Paris StreetView, CelebA, and Place2 datasets. Fig. \ref{fig:Paris}, ~\ref{fig:CelebA} and ~\ref{fig:Place2} show three corrupted images, their groundtruth, the inpainted images by the proposed and benchmark algorithms. The proposed algorithm can generate realistic details and structures. Specifically, in the top row of Fig. \ref{fig:Paris}, the window produced by the proposed algorithm is clearer than those produced by other methods. Further evidence could be seen in the top row of Fig. \ref{fig:CelebA} and both rows of Fig. \ref{fig:Place2}. In the former, the hair strands atop the man's forehead are better defined and clearer in the image produced by the proposed algorithm compared to those produced by the benchmark algorithms; they look similar to those strands depicted in the groundtruth image. Both rows of Fig.\ref{fig:Place2} show artifacts in the inpainted region by benchmark algorithms while the proposed algorithm does not suffer from such artifacts, producing a more pleasing image to the eye.

\begin{table*}[!htbp]
	\centering
	\begin{threeparttable}
		\centering
		\caption{Ablation study results on the Paris Streetview Dataset based on network structure}
		\begin{tabular}{c|c|c|c|c|c|c|c|c|c|c|c}
			\hline
			\multicolumn{3}{c|}{Ablation setting} & \multicolumn{3}{c|}{PSNR} & \multicolumn{3}{c|}{SSIM} & \multicolumn{3}{c}{Mean $l_1$}\\
			\cline{1-12}
			{Reinpainting-1} & {Reinpainting-2} & {GLE} & 10-20\%  & 30-40\%  & 40-50\%  & 10-20\%  & 30-40\%  & 40-50\%  & 10-20\%  & 30-40\%  & 40-50\%  \\
			\cline{1-12}
			\XSolidBrush & \XSolidBrush & \XSolidBrush & {31.82} & {26.70} & {24.83} & {0.9632} & {0.8858} & {0.8276} & {0.0074} & {0.0196} & {0.0277}\\
			\XSolidBrush & \XSolidBrush & \Checkmark & {31.90} & {26.76} & {24.86} & {0.9637} & {0.8859} & {0.8285} & {0.0073} & {0.0195} & {0.0275}\\
			\XSolidBrush & \Checkmark & \XSolidBrush & {31.92} & {26.83} & {24.93} & {0.9638} & {0.8886} & {0.8314} & {0.0073} & {0.0193} & {0.0272}\\
			\Checkmark & \XSolidBrush & \XSolidBrush & {31.93} & {26.83} & {24.92} & {0.9640} & {0.8887} & {0.8312} & {0.0073} & {0.0192} & {0.0272}\\
			\Checkmark & \XSolidBrush & \Checkmark & {31.95} & {26.84} & {24.96} & {0.9641} & {0.8888} & {0.8317} & {0.0073} & {0.0192} & {0.0271}\\
			\XSolidBrush & \Checkmark & \Checkmark & {31.99} & {26.86} & {25.04} & {0.9642} & {0.8890} & {0.8330} & {0.0073} & {0.0192} & {0.0268}\\
			\cline{1-12}
		\end{tabular}
		\label{Tab:Ablation}
	\end{threeparttable}
\end{table*}

\begin{table*}[!htbp]
	\centering
	\begin{threeparttable}
		\centering
		\caption{Ablation study results on the Paris Streetview Dataset based on loss function}
		\begin{tabular}{c|c|c|c|c|c|c|c|c|c|c|c}
			\hline
			\multicolumn{3}{c|}{Ablation setting} & \multicolumn{3}{c|}{PSNR} & \multicolumn{3}{c|}{SSIM} & \multicolumn{3}{c}{Mean $l_1$}\\
			\cline{1-12}
			{$L_{valid}$ \& $L_{hole}$} & {$L_{perc}$ \& $L_{style}$} &  {$L_{tv}$}  & 10-20\%  & 30-40\%  & 40-50\%  & 10-20\%  & 30-40\%  & 40-50\%  & 10-20\%  & 30-40\%  & 40-50\%  \\
			\cline{1-12}
			\Checkmark & \XSolidBrush & \XSolidBrush &{31.72} & {26.61} & {24.85} & {0.9604} & {0.8817} & {0.8237} & {0.0072} & {0.0193} & {0.0270}\\
			\XSolidBrush & \Checkmark & \XSolidBrush &{31.69} & {26.72} & {24.96} & {0.9599} & {0.8830} & {0.8254} & {0.0079} & {0.0200} & {0.0278}\\
			\Checkmark & \Checkmark & \XSolidBrush &{31.86} & {26.70} & {24.92} & {0.9616} & {0.8823} & {0.8223} & {0.0076} & {0.0199} & {0.0279}\\
			\Checkmark & \Checkmark & \Checkmark & {31.99} & {26.86} & {25.04} & {0.9642} & {0.8890} & {0.8330} & {0.0073} & {0.0192} & {0.0268}\\
			\cline{1-12}
		\end{tabular}
		\label{Tab:Ablation-loss}
	\end{threeparttable}
\end{table*}

\subsection{Ablation Analysis on Proposed Architecture}
The proposed architecture is closely examined to get a better understanding of the role of the novel components. More to the point, the GLE module and the reinpainting component are novel parts that set the proposed architecture apart form the other inpainting algorithms. Therefore, this section will focus on shedding some light on their roles in the inpainting process. The objective is to address the question: \textit{how much of an impact do the GLE module and the reinpainting component have on the performance of the algorithm?} This is going to be done in three experiments. The first has the two parts removed and the performance of the remaining architecture is evaluated. This helps establish the baseline results. The other two experiments examine the impact of adding each of the two parts, i.e., GLE and reinpainting, separately on the inpainting performance. The results of the five experiments are shown below.

\subsubsection{Removing the GLE Module and Reinpainting Component}\label{sec:abl_exp_1}
The GLE module and reinpainting component are both removed from the proposed architecture. To avoid jeopardizing the capacity of the proposed model, the GLE module is removed by stripping away the Gaussian smoothing and upsampling layers making a direct path from the first to the second convolution layers of the module.

Removing both parts chips away from the inpainting performance of the architecture. This is evident in Table~\ref{Tab:Ablation}; with all three metrics, the table shows a clear degradation in performance on the Paris StreetView dataset when the architecture is trained and tested without the GLE module and the reinpainting component.

\subsubsection{Removing the GLE Module}
Using the same removal strategy in the above section (Section \ref{sec:abl_exp_1}), the GLE module is removed in this experiment while keeping the reinpainting component. The result of doing so is a slight improvement in the performance compared to the baseline case, i.e., no GLE and reinpainting, as Table~\ref{Tab:Ablation} shows. However, the performance is still worse compared to having both parts plugged in. The results of this experiment could be used to argue for the value of the GLE module; it helps the proposed architecture extract expressive features from different frequency components of the image.

\subsubsection{Removing the Reinpainting Component}
Using the same removing strategy once again, the reinpainting component is removed while keeping the GLE module. This setting is labelled as Reinpainting-1 in Table~\ref{Tab:Ablation}. It is hypothesized that reinpainting has the ability to fill large holes by accessing features from neighboring iterations. The results in Table~\ref{Tab:Ablation} verify that hypothesis to some extent; removing the reinpainting component degrades the performance of the architecture despite the presence of the GLE module.

\subsubsection{Value of progression for reinpainting}
The hypothesis about the reinpainting component being able to fill large holes is further examined here. More to the point, it will be argued that accessing sub-volumes from different iterations (i.e., $\mathbf{F}_{\text{int}}(\tau-1)$ and $\mathbf{F}_{\text{int}}(\tau+1)$) has added value to the inpainting process. This is first done by restricting the input to the re-inpainting component to only the $\tau$-th feature sub-volume, i.e., $\mathbf{F}_{\text{int}}(\tau)$. Then, two experiments are conducted with and without the GLE components.
\begin{itemize}
	\item The GLE modules are removed as described in Section \ref{sec:abl_exp_1}. The performance in this case is very close to that of removing the whole reinpainting component. This is indicated in Table~\ref{Tab:Ablation} under Reinpainting-2. This verifies that the $\mathbf{F}_{\text{int}}(\tau)$ and $\mathbf{\tilde H} (\tau)$ can not provide more useful information for the reinpainting process. The redundant information even sightly causes the performance to degrade.
	\item The GLE modules are put back and the experiment is repeated again. Again, the results, shown in Table~\ref{Tab:Ablation}, further verify that the input features from neighbouring iterations are useful for enhancing the re-inpainting results.
\end{itemize}

\subsection{Ablation Analysis on Loss Function}
The performance of the proposed method is further investigated based on each component of the loss function. The ablation analysis is done in three steps, each of which is illustrating the incremental value of certain terms in the loss function. The three steps are discussed below:
\begin{itemize}
	\item \textbf{Using $L_{valid}$ and $L_{hole}$:} Using $L_{valid}$ and $L_{hole}$ as the only terms of the loss functions, we observe that the mean $l_1$ is slightly better compared to the final results; however, the PSNR and SSIM are not satisfactory especially for the case of corrupted images with large holes. The undesirable performance indicates that the loss has to account for the semantic information in the inpainted image.
	\item \textbf{Using $L_{perc}$ and $L_{style}$:} Constructing the loss function using $L_{perc}$ and $L_{style}$ alone, we observe an improvement in the PSNR and SSIM for large holes. The performance shows the role and effectiveness of $L_{perc}$ and $L_{style}$ in discovering semantic information in the valid region; however, the mean $l_1$ loss increases significantly due to the lack of $L_{valid}$ and $L_{hole}$.
	\item \textbf{Using $L_{valid}$, $L_{hole}$, $L_{perc}$ and $L_{style}$:} After using the $L_{valid}$, $L_{hole}$, $L_{perc}$ and $L_{style}$ as the terms of the loss function, the PSNR, SSIM and mean $l_1$ all improve but for the case of low mask ratio. However, for the case of high mask ratio, the PSNR and SSIM resulting from using $L_{perc}$ and $L_{style}$ alone are slightly better than the case of combining all four $L_{valid}$, $L_{hole}$, $L_{perc}$ and $L_{style}$. It is hypothesized that the noise may be generated from the pixel and feature levels. To eliminate such noise, the $L_{tv}$ is added to the loss function, which improves the overall performance.
\end{itemize}

\section{Conclusion}
This paper introduces a three-stage neural network architecture that is able to progressively inpaint corrupted images while maintaining their structural and contextual integrity. In its core is a novel Gaussian-Laplacian feature Extraction (GLE) module. Stacking GLE modules constructs the first stage of the architecture and enables the network to build a feature pyramid of different frequency components, disintegrating structural (high frequency) and contextual (low frequency) information. The feature pyramid is the key for structurally- and contextually-aware progressive inpainting; low- and high-frequency components are iteratively but separately inpainted and fused in the second stage. The third, and final, stage enhances the fused features before it reconstructs the inpainted image. Experimental results and benchmarking show that the three-stage architecture is able to restore fine details in the corrupted region, outperforming the state-of-the-art algorithms. Ablation experiments reveal that the GLE module and the reinpainting component are responsible for the superior performance of the proposed architecture.


\bibliographystyle{IEEEtran}
\bibliography{reference}

\end{document}